\documentclass{article}

\usepackage[preprint,nonatbib]{neurips_2021}

\usepackage[utf8]{inputenc} % allow utf-8 input
\usepackage[T1]{fontenc}    % use 8-bit T1 fonts
\usepackage{hyperref}       % hyperlinks
\usepackage{url}            % simple URL typesetting
\usepackage{graphicx}
\usepackage{booktabs}       % professional-quality tables
\usepackage{amsfonts}       % blackboard math symbols
\usepackage{nicefrac}       % compact symbols for 1/2, etc.
\usepackage{microtype}      % microtypography
\usepackage{xcolor}         % colors
\usepackage{subcaption}

\title{LAION-400M: Open Dataset of CLIP-Filtered 400 Million Image-Text Pairs}

\author{Christoph Schuhmann \\
LAION \\
\texttt{contact@laion.ai} \\
\And
Richard Vencu \\
LAION \\
Gentec Data \\
\texttt{richard.vencu@gentec.ro} \\
\And
Romain Beaumont \\
LAION \\
\texttt{romain.rom1@gmail.com} \\
\And
Robert Kaczmarczyk\\
LAION\\
Technical University of Munich\\
\texttt{robert.kaczmarczyk@tum.de} \\
\And
Clayton Mullis \\
LAION\\
\texttt{claymullis@fastmail.com} \\
\And
Aarush Katta \\
LAION \\
\texttt{ARKsealplays@gmail.com} \\
\And
Theo Coombes \\
LAION \\
\texttt{theocoombes06@gmail.com} \\
\And
Jenia Jitsev \\
LAION \\
Juelich Supercomputing Center (JSC)\\ 
Research Center Juelich (FZJ) \\
\texttt{j.jitsev@fz-juelich.de}\\
\And
Aran Komatsuzaki \\
LAION \\
Georgia Institute of Technology\\
EleutherAI\\
\texttt{akomatsuzaki3@gatech.edu} \\
}

\begin{document}

\maketitle

\begin{abstract}
Multi-modal language-vision models trained on hundreds of millions of image-text pairs (e.g. CLIP, DALL-E) gained a recent surge, showing remarkable capability to perform zero- or few-shot learning and transfer even in absence of per-sample labels on target image data. Despite this trend, to date there has been no publicly available datasets of sufficient scale for training such models from scratch. To address this issue, in a community effort we build and release for public LAION-400M, a dataset with CLIP-filtered 400 million image-text pairs, their CLIP embeddings and kNN indices that allow efficient similarity search.\footnote{Project page: \href{https://laion.ai/laion-400-open-dataset/}{https://laion.ai/laion-400-open-dataset/}} 
\end{abstract}

\section{Introduction}
\label{intro}

% \subsection{Background}

% Multi-modal models are important especially for data efficient learning and transfer to novel datasets. However, this ability can be expressed only when employing sufficiently large scales, both for model and data during training. Increasing scale is known to improve generalization and transfer from works in language and vision modelling. ... So, without large scale data, no strong transfer possible, no proper studies of such models possible.
% It even prevents the saturation of performance improvement when the compute budget is increased, which allows models to be scaled up more efficiently

% \textbf{Background.} 
Multi-modal language-vision models demonstrated recently strong transfer capability to novel datasets in absense of per-sample labels~\cite{clip, dalle, align}. This capability requires sufficiently large model and data scale during pre-training. Increasing data scale alone can often improve model performance \cite{epoch}. When increasing model and compute budget scale in addition, scaling laws suggest further increase in generalization and transfer performance if not bottlenecked by the data scale \cite{kaplan, kaplan2, Kolesnikov2020, zhai2021scaling}. There is a plethora of recent works that have built massive datasets in order to optimally scale up various models \cite{gpt3, clip, dalle, align}. However, these massive datasets have rarely been released for various reasons. Gao et. al. recently released The Pile, an openly-available 800GB text dataset~\cite{pile}, in an attempt to loosely mimic the dataset used for GPT-3. The largest publicly known image-text paired datasets range from 400 million to around a billion, but none of them has been released. 

To address this issue, we build and release LAION-400M, a dataset with CLIP-filtered 400 million image-text pairs, their CLIP embeddings and kNN indices. We describe the procedure to create the dataset and demonstrate successful training of DALL-E architecture. Having sufficiently large scale, the dataset opens venues for research on multi-modal language-vision models to broad community.

% The openly availaible dataset has scale that is large enough to allow broad research on multi-modal language-vision models

\section{Dataset and Methods}

% width=0.7\linewidth

\begin{figure}[!tb]
    \centering
    \subfloat{{\includegraphics[width=\linewidth]{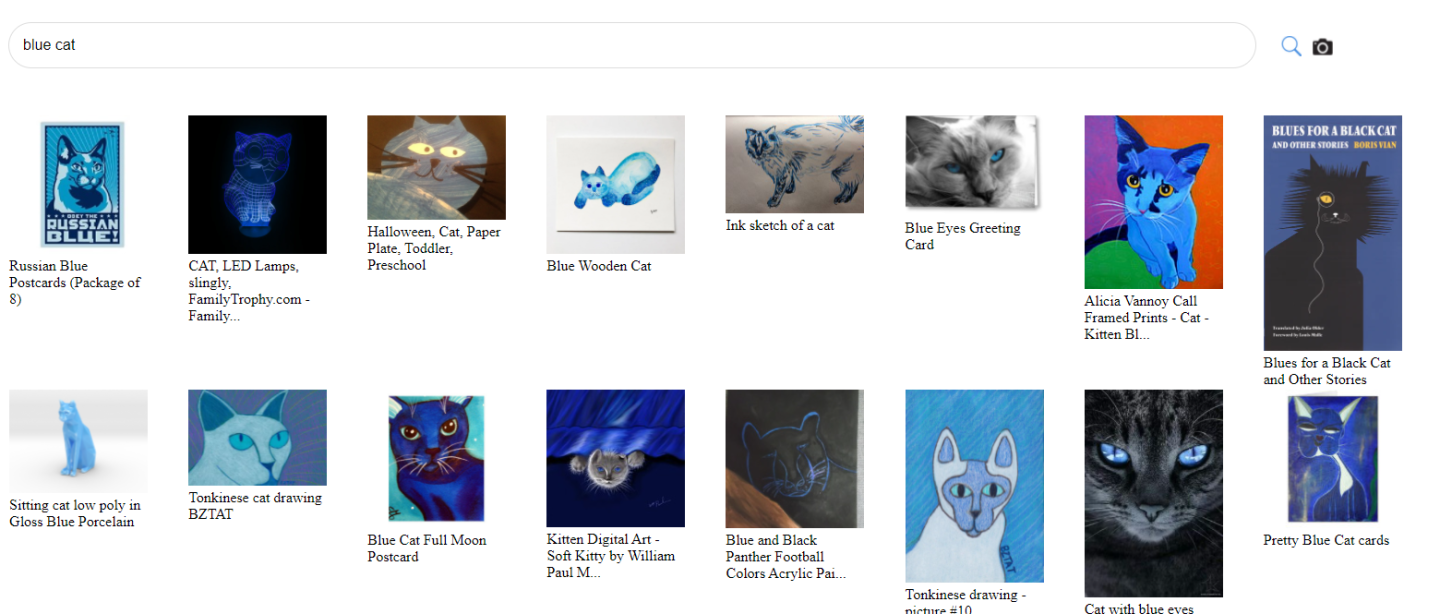}}}\\
    \subfloat{{\includegraphics[width=\linewidth]{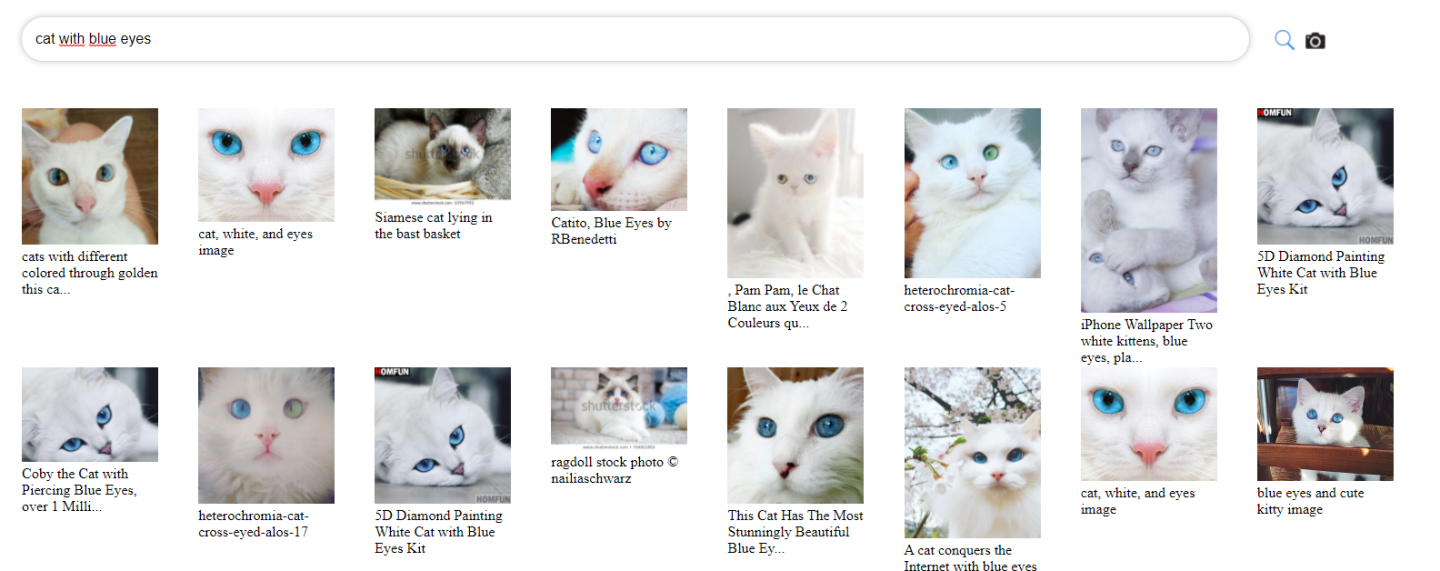}}}\\
    \caption{Sample images retrieved from the queries "blue cat" or "cat with blue eyes" in the web demo}
    \label{sample}
\end{figure}

\begin{figure}[!t]
    \centering
    \includegraphics[scale=0.5]{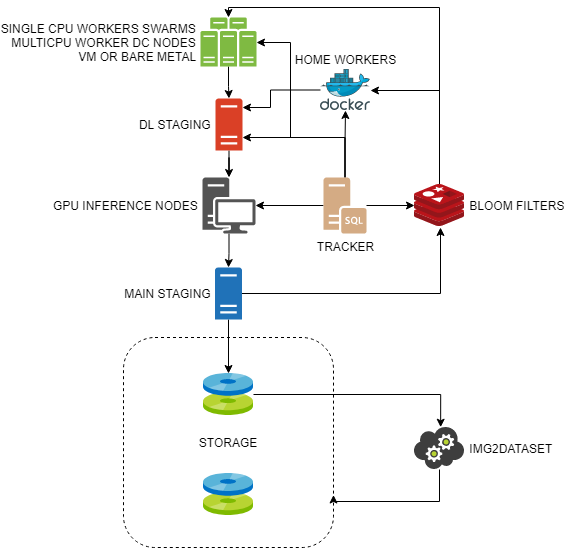}
    \caption{Acquisition workflow}
    \label{arch}
\end{figure}

% \subsection{Overview of LAION-400M}
\textbf{Overview of LAION-400M.} We officially release the following packages under LAION-400M project: 
\begin{itemize}
    \item 400 million pairs of image URL and the corresponding metadata
    \item 400 million pairs of CLIP image embedding and the corresponding text
    \item Several sets of kNN indices that enable quick search in the dataset
    \item img2dataset library that enables efficient crawling and processing of hundreds of millions of images and their metadata from a list of URLs with minimal resources
    \item Web demo of image-text search on LAION-400M (Fig. \ref{sample})\footnote{\href{https://rom1504.github.io/clip-retrieval/}{https://rom1504.github.io/clip-retrieval/}}
\end{itemize}

As for the pairs of image URL and metadata, we provide parquet files that consist of the following attributes for each pair: sample ID, URL, type of Creative Commons license (if applicable), NSFW tag (detected with CLIP), cosine similarity score between the text and image embedding and height and width of the image. We found less than 1\% of images were detected as NSFW, which can be filtered out by an user with NSFW tag.

%\begin{itemize}
%    \item 400 million pairs of image URL and the corresponding metadata 
%    \item 400 million pairs of CLIP image embedding and the corresponding text
%    \item Several sets of kNN indices that enable quick search in the dataset
%    \item img2dataset library: 
%\end{itemize}

\textbf{Acquisition.} The acquisition follows the flowchart of Fig. \ref{arch} and can be split into two major components: 
\begin{itemize}
    \item Distributed processing of petabyte-scale Common Crawl dataset, which produces a collection of matching URLs and captions.
    \item Single node post-processing of the data, which is much lighter and can be run in a few days, producing the final dataset.
\end{itemize}

%\begin{itemize}
%    \item Distributed processing of the huge (many PBs) Common Crawl dataset, which produces a collection of matching URLs and captions.
%    \item Single node much lighter post-processing of the data, that anyone can run in a few days and which produces the final dataset.
%\end{itemize}

\subsection{Distributed processing of Common Crawl}
To create image-text pairs, we parse through WAT files from Common Crawl and parse out all HTML IMG tags containing an alt-text attribute. We download the raw images from the parsed URLs with asynchronous requests using Trio and Asks libraries. 

\subsubsection{Filtering out unsuitable image-text pairs}
After downloading the WAT files from Common Crawl, we apply the following filtering conditions:
\begin{itemize}
    \item All samples with less than 5 character alt-text length or less than 5 KB image size are dropped.
    \item Duplicate removal is performed with bloom filter based on URL and alt-text.
    \item We use CLIP to compute embeddings of the image and alt-text. Then we compute the cosine similarity of both embeddings and drop all samples with cosine similarity below 0.3. This threshold was selected based on human inspections. 
    \item We use the CLIP embeddings of images and texts to filter out illegal contents. 
\end{itemize}

\subsubsection{img2dataset}
We developed img2dataset library to comfortably download from a given set of URLs, resize and store the images and captions in the webdataset format.\footnote{\href{https://github.com/rom1504/img2dataset}{https://github.com/rom1504/img2dataset}} This allows to download 100 million images from our list of URLs in 20 hours with a single node (1Gbps connection speed, 32GB of RAM, an i7 CPU with 16 cores), which allows anyone to obtain the whole dataset or a smaller subset.

\section{Analysis \& Results}

\textbf{Web demo and similarity search.} A web demo was created to allow an user to search images and texts based on a query image or text using the CLIP embeddings of the input and our precomputed kNN indices. It demonstrates the diversity of images and captions that can be found in LAION-400M as well as high semantic relevance (Fig. \ref{sample}).

Tab. \ref{table} shows the distribution of image sizes of LAION-400M. Given the abundance of high-resolution images, one can produce subsets of images for training various customized models, and also choose image resolution that is suitable for purpose of particular training. 

\begin{table}[!tb]
\begin{center}
\begin{tabular}{||c c||} 
 \hline
 Number of unique samples & 413M  \\ 
 \hline
 Number with height or width $\geq$ 1024 & 26M \\
 \hline
 Number with height and width $\geq$ 1024 & 9.6M \\
  \hline
 Number with height and width $\geq$ 512 & 67M \\
 \hline
 Number with height or width $\geq$ 512 & 112M \\
  \hline
Number with height and width $\geq$ 256 & 211M \\
 \hline
Number with height or width $\geq$ 256 & 268M \\
 [1ex] 
 \hline
\end{tabular}
\end{center}
\caption{Image size distribution of LAION-400M}
\label{table}
\end{table}

\begin{figure}[!tb]
\begin{subfigure}{.55\textwidth}
  \centering
  \includegraphics[width=\linewidth]{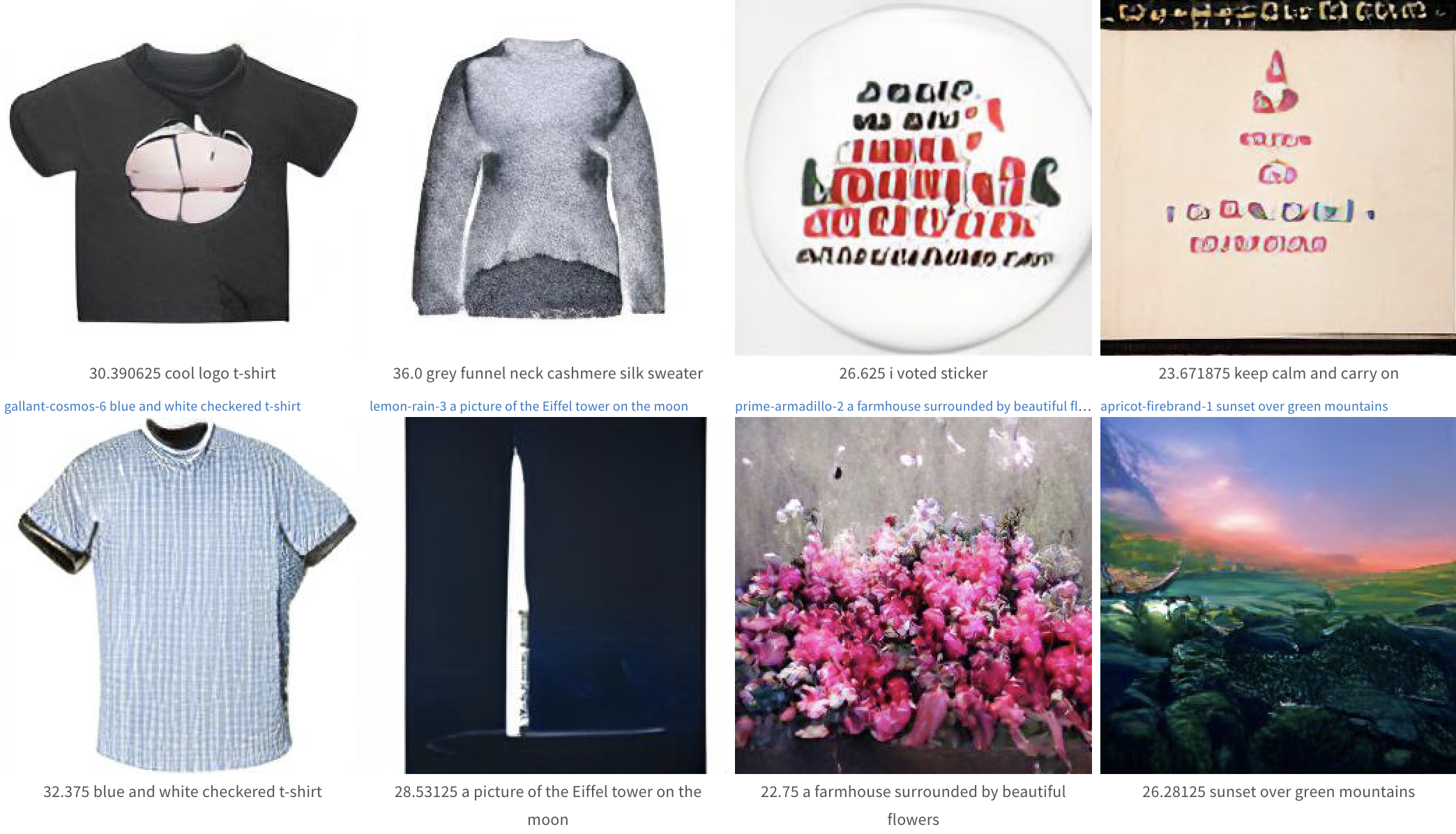}
  %\caption{Put your sub-caption here}
  %\caption{}
  \label{fig:dalle_samples}
\end{subfigure}
\begin{subfigure}{.45\textwidth}
  \centering
  % include second image
  \includegraphics[width=\linewidth]{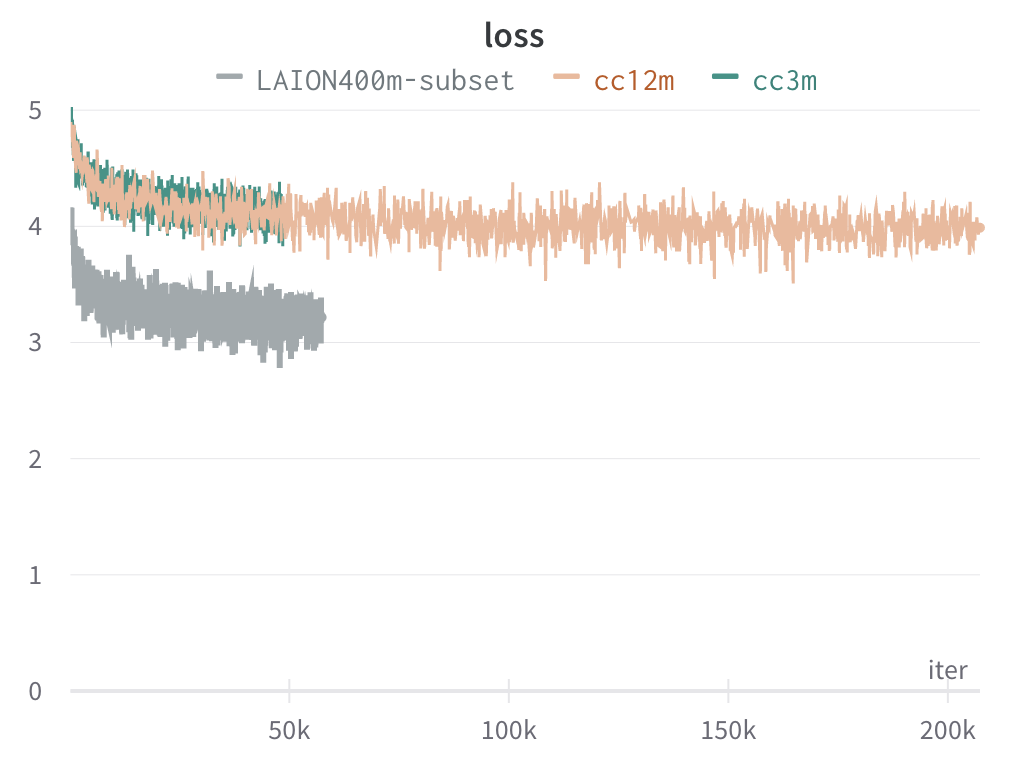}
  %\caption{Put your sub-caption here}
  %\caption{}
  \label{fig:dalle_comparison}
\end{subfigure}
\vspace*{-0.3cm}
\caption{DALL-E Experiments. \textbf{(Left)} Generated samples from a DALL-E model trained with 7.2M randomly picked LAION-400M samples on 1 RTX 2070 Super (8 GB VRAM) for 1 epoch \textbf{(Right)} DALL-E runs with Conceptual Captions 3M (green), Conceptual Captions 12M (orange) and a 3M subset of LAION-400M (grey)}
\label{fig:dalle_exp}
\end{figure}

% \subsection{Analysis}

% \subsection{Training DALL-E}
\textbf{Training DALL-E model.} We ran DALLE-pytorch \cite{dallepytorch2021}, an open-source replication of DALL-E \cite{dalle}, to assess the dataset's capability to train a text-to-image model. The VQGAN \cite{vqgan} pretrained on ImageNet is used to encode image tokens. For generation, we use CLIP ViT-B/16 \cite{clip} to rank the top 8 of 128 total samples per caption. Despite only seeing a subset of approximately 7.2 million images for a single epoch, we observe fast convergence across a variety of categories. Samples generated from the model show sufficient quality and provide evidence for successful training progress (Fig. \ref{fig:dalle_exp}).

\section{Conclusion}
By releasing an openly available dataset that contains 400 million image-text pairs, we have closed the gap to proprietary large scale datasets that were necessary to train state-of-the-art language-vision models such as DALL-E and CLIP. As proof of concept, we demonstrated that a subset of our dataset can be used to train a DALL-E model, producing samples of sufficient quality. The dataset opens the road for large-scale training and research of language-vision models, that were previously restricted to those having access to proprietary large datasets, to the broad community. 

% and providing evidence of dataset's suitability
% It shows

%\bibliographystyle{plainnat}
% to make citations in the text shorter
\bibliographystyle{unsrt}
\bibliography{bibfile}

\end{document}